\documentclass[journal]{IEEEtran}
\usepackage{cite}
\usepackage{graphicx}
\graphicspath{{./figures/}}
\usepackage{amsmath}
\usepackage[hypertexnames=false,colorlinks=true,linkcolor=black,%
 anchorcolor=black,citecolor=black,urlcolor=black]{hyperref}
\usepackage{fixltx2e}
\usepackage{url}
\usepackage{microtype,amsfonts,subfig,xcolor}
\usepackage{mathptmx}
\usepackage[scaled=.9]{helvet}

\newcommand{\tbname}[1]{\textbf{\textsf{#1}}}
\newcommand{\colname}[1]{\textsf{#1}}
\newcommand*\circled[1]{\raisebox{.5pt}{\textcircled{\raisebox{-.9pt}{\small\textsf{#1}}}}}

\begin{document}
\title{An Extensible Benchmarking Infrastructure\\ for Motion Planning Algorithms}
\author{Mark Moll, Ioan A. \c Sucan, and Lydia E.~Kavraki%
\thanks{M.~Moll and L.E.~Kavraki are with the Department of Computer Science at Rice University, Houston, TX 77005, USA. Email: \{mmoll,kavraki\}@rice.edu. I.A.~\c Sucan is with Google[X], Mountain View, CA, USA. Email: isucan@gmail.com.}}
\maketitle
\footernote{Submitted to:
IEEE Robotics \& Automation Magazine (Special Issue on Replicable and Measurable Robotics Research), 2015}

\section{Introduction}

\IEEEPARstart{M}{otion} planning is a key problem in robotics concerned with finding a path that satisfies a goal specification subject to constraints. In its simplest form, the solution to this problem consists of finding a path connecting two states and the only constraint is to avoid collisions. Even for this version of the motion planning problem there is no efficient solution for the general case~\cite{canny1988:compl-robot-motion-plann}. The addition of differential constraints on robot motion or more general goal specifications make motion planning even harder. Given its complexity, most planning algorithms forego completeness and optimality for slightly weaker notions such as \emph{resolution} completeness or \emph{probabilistic} completeness~\cite{choset2005principles-of-robot-motion:-theory} and \emph{asymptotic} optimality. 

\begin{figure}
  \centering\includegraphics[width=.7\linewidth]{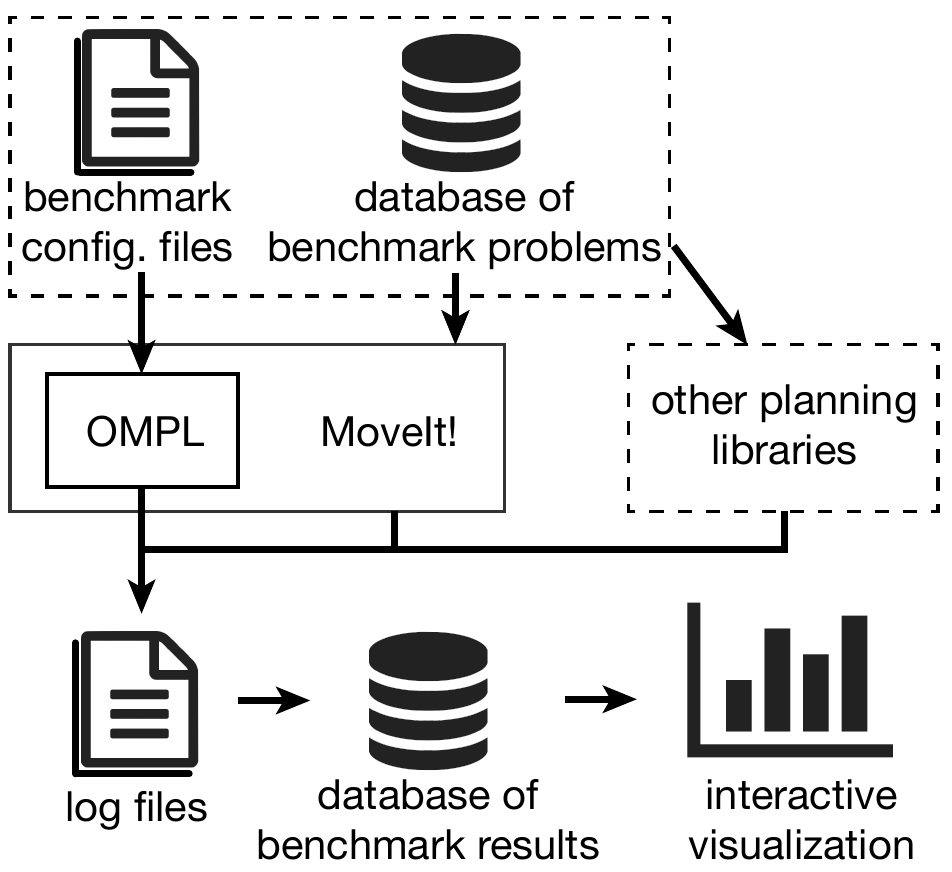}
  \caption{Overview of the benchmarking infrastructure.}
  \label{fig:overview}
\end{figure}

Sampling-based planning algorithms are the most common probabilistically complete algorithms and are widely used on many robot platforms. Within this class of algorithms, many variants have been proposed over the last 20 years, yet there is still no characterization of which algorithms are well-suited for which classes of problems. This has motivated us to develop a benchmarking infrastructure for motion planning algorithms (see Figure~\ref{fig:overview}). It consists of three main components. First, we have created an extensive benchmarking software framework that is included with the Open Motion Planning Library (OMPL), a C++ library that contains implementations of many sampling-based algorithms~\cite{sucan2012the-open-motion-planning-library}.
One can immediately compare any new planning algorithm to the 29 other planning algorithms that currently exist within OMPL. There is also much flexibility in the types of motion planning problems that can be benchmarked, as discussed in Section~\ref{sec:def-motion-planning-problems}.
Second, we have defined extensible formats for storing benchmark results. The formats are fairly straightforward so that other planning libraries could easily produce compatible output. Finally, we have created an interactive, versatile visualization tool for compact presentation of collected benchmark data. The tool and underlying database facilitate the analysis of performance across benchmark problems and planners. While the three components described above emphasize generality, we have also created---as an example---a simple command line tool specifically for rigid body motion planning that takes as input a plain text description of a motion planning problem.

Benchmarking sampling-based planners is non-trivial for several reasons. Since these planners rely on sampling, performance cannot be judged from a single run. Instead, benchmarks need to be run repeatedly to obtain a \emph{distribution} of some performance metric of interest. Simply comparing the means of such distributions may not always be the correct way to assess performance. Second, it is well-known that different sampling strategies employed by sampling-based algorithms typically perform well only for certain classes of problems, but it is difficult to exactly define such classes. Finally, different applications require optimization for different metrics (e.g., path quality versus time of computation) and there is no universal metric to assess performance of planning algorithms across all benchmarks.



There have been some attempts in the past to come up with a general infrastructure for comparing different planning algorithms (see, e.g., \cite{gipson2001mpk:-an-open-extensible-motion, cohen2012a-generic-infrastructure-for-benchmarking-motion}). Our work provides similar benchmarking capabilities, but includes an extended and extensible set of metrics, offers higher levels of abstraction and at the same time concrete entry level points for end users. 
Furthermore, we also introduce an extensible logging format that other software can use and a visualization tool. To the best of the authors' knowledge, none of the prior work offered the ability to interactively explore and visualize benchmark results.

\section{Benchmarking infrastructure}

OMPL provides a high-level of abstraction for defining motion planning problems. The planning algorithms in OMPL are to a large extent agnostic with respect to the space they are planning in. Similarly, the benchmarking infrastructure within OMPL allows the user to collect various statistics for different types of motion planning problems. The basic workflow is as follows:
\begin{enumerate}
  \item The user defines a motion planning problem. This involves defining the state space of the robot, a function which determines which states are valid (e.g., collision-free), the start state of the robot and the goal. The complete definition of a motion planning problem is contained within a C++ object, which is used to construct a benchmark object.
  \item The user specifies which planning algorithms should be used to solve the problem, time and memory limits for each run, and the number of runs for each planner.
  \item The benchmark is run. Upon completion, the collected results are saved to a log file. A script is used to add the results in the log file to an SQL database. The results can be queried directly in the database, or explored and visualized interactively through a web site set up for this purpose (\url{http://plannerarena.org}).
\end{enumerate}
Below we will discuss these steps in more detail.

\subsection{Defining motion planning problems}
\label{sec:def-motion-planning-problems}
The most common benchmark motion planning problems are rigid body problems, due to their simplicity (it is easy for users to intuitively assess performance). We have developed a simple plain-text file format that describes such problems with a number of key-value pairs. Robots and environments are specified by mesh files. The state validity function is in this case hard-coded to be a collision checker. Besides the start and goal positions of the robot, the user can also specify an \emph{optimization objective}: path length, minimum clearance along the path, or mechanical work. There are several planning algorithms in OMPL that optimize a path with respect to a specified objective. (Others that do not support optimization simply ignore this objective.) It is also possible to specify simple kinodynamic motion planning problems. OMPL.app, the application layer on top of the core OMPL library,  predefines the following systems that can be used: a first-order car, a second-order car, a blimp, and a quadrotor. 
We have not developed controllers or steering functions for these systems and kinodynamic planners in OMPL fall back in such cases on sampling random controls. This makes planning for these systems extremely challenging. With a few lines of code, the command line tool can be modified to allow new planning algorithms or new types of planning problems to be specified in the configuration files.

The benchmark configuration files can be created with the GUI included with OMPL.app. A user can load meshes in a large variety of formats, define start and goal states, try to solve the problem with different planners and save the configuration file. The user can also visualize the tree/graph produced by a planning algorithm to get a sense of how hard a particular problem is. In the configuration file, the user can specify whether solution paths (all or just the best one) should be saved during benchmarking. Saved paths can be ``played back'' with the GUI.

When defining motion planning problems in code, many of the limitations of the command line tool go away. Arbitrary state spaces and kinodynamic systems can be used, different notions of state validity can be defined, and different optimization objectives can be defined. Additionally, any user-defined planning algorithm can be used. The OMPL application programmer interface (API) imposes only minimal requirements on new planning algorithms. In particular, the API is not limited to sampling-based algorithms (in \cite{luo2014an-empirical-study-of-optimal-motion}, for example, several non-sampling-based planners are integrated into OMPL). The low barrier to entry has lead to numerous contributions of planning algorithms from other groups: OMPL~1.0 includes 29 planning algorithms. Since all these algorithms use the same low-level functionality for, e.g., collision checking, benchmarking highlights the differences in the motion planning algorithms themselves.

The benchmarking facilities in MoveIt!~\cite{sucan-moveit} are based on and compatible with those in OMPL. The problem setup is somewhat similar to the OMPL command line tool. In MoveIt!, robots are specified by URDF files, which specify a robot's geometry and kinematics. Motion planning problems to be benchmarked are stored in a database.

\subsection{Specifying planning algorithms}

Once a motion planning problem has been specified, the next step is to select one or more planners that are appropriate for the given problem. Within OMPL, planners are divided into two categories: geometric/kinematic planners and kinodynamic planners. The first category can be further divided into two subcategories: planners that terminate when \emph{any} solution is found and planners that attempt to compute an \emph{optimal} solution (with respect to a user-specified optimization objective). For optimizing planners a threshold on optimality can be set to control how close to optimal the solution needs to be. At one extreme, when this threshold is set to 0, planners will run until time runs out. At the other extreme, when the threshold is set to infinity, planners act like the non-optimizing planners and will terminate as soon as any solution is found.

Typically, a user specifies multiple planners. By default, OMPL will try to make reasonable parameter choices for each planner. However, a user can also fine-tune any parameter setting for a planner. With the command line tool's configuration files, this is easily accomplished by adding lines of the form ``\texttt{planner.parameter=value}.''
The parameter code infrastructure is generic: when a programmer specifies a parameter for a planner, it can be specified through the configuration file without having to change the parsing of configuration files.
It is also possible to add many instances of the same type of planner. This is useful for, e.g., parameter sweeps. Each instance can be given a slightly different name to help distinguish the results for each instance. Each run of a planner is run in a separate thread, so that if a planner hangs, the benchmark program can detect that and forcibly terminate the planner thread (the run is recorded as a crash and the benchmarking will continue with the next run).

\begin{figure}
  \centering\includegraphics[width=.85\linewidth]{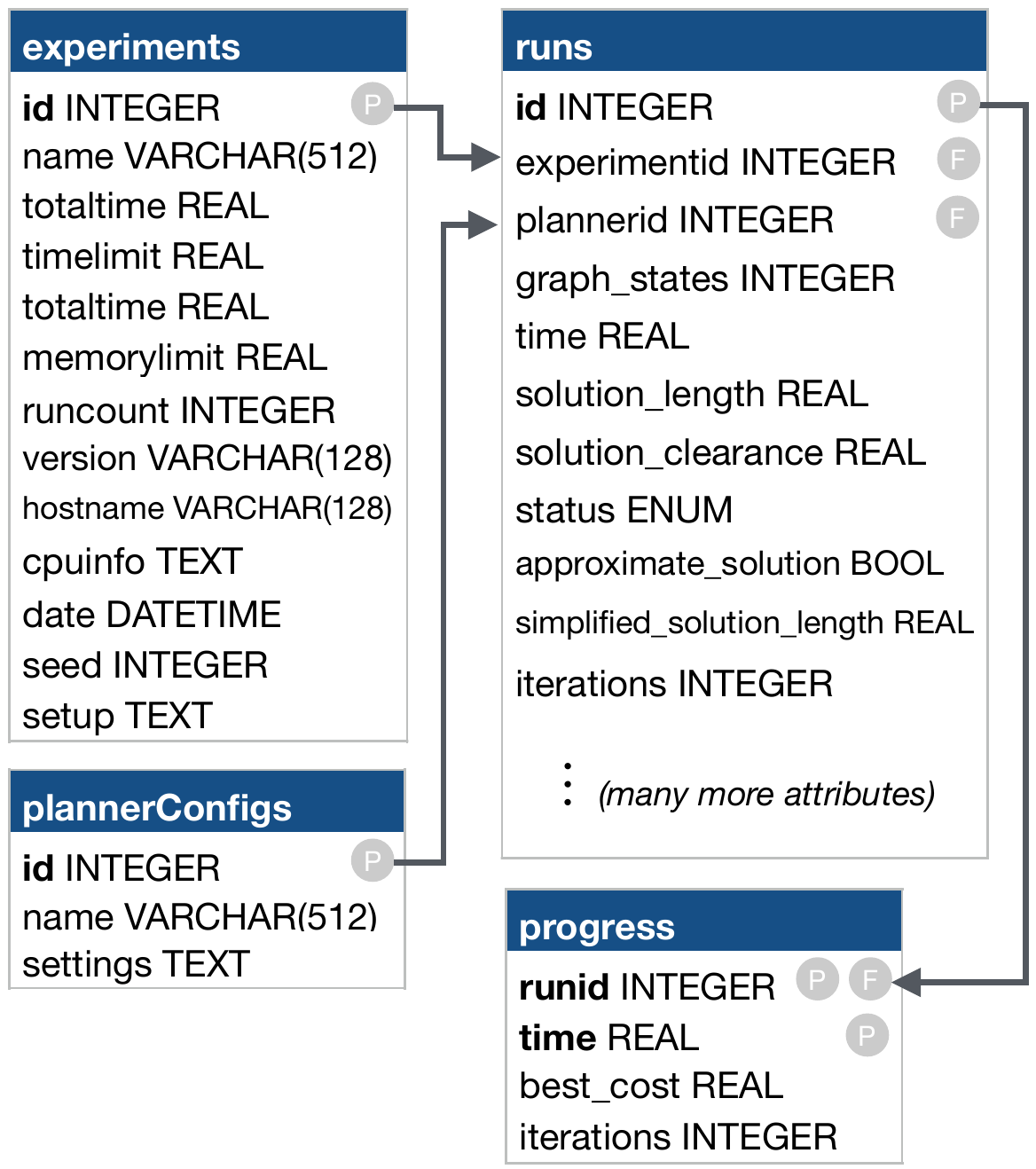}
  \caption{Schema for a database of benchmark results. The \circled{P} and \circled{F} denote the primary and foreign keys of each table, respectively.}
  \label{fig:dbschema}
\end{figure}

\subsection{A database of benchmark runs}

After a benchmark run is completed, a log file is written out. With the help of a script, the benchmark results stored in the log file can be added to a SQLite3 database. Multiple benchmark log files can be added to the same database. The SQLite3 database facilitates distribution of all relevant benchmark data: users can simply transfer one single file. Furthermore, the database can be easily programmatically queried with almost any programming language.

Figure~\ref{fig:dbschema} illustrates the database schema that is used. Each benchmark log file corresponds to one experiment. The \tbname{experiments} table contains an entry for each experiment that contains the basic benchmark parameters, but also detailed information about the hardware on which the experiment was performed (in the \colname{cpuinfo} column). Information about each of the planner instances that were specified is stored in the \tbname{plannerConfigs} table. For each planner instance, all parameter values are stored as a string representation of a list of key-value pairs (in the \colname{settings} column). While we could have created a separate column in the \tbname{plannerConfigs} table for each parameter, the parameters are very planner specific with very few shared parameters among planners.


The main results are stored in the \tbname{runs} table. Each entry in this table corresponds to one run of a particular planner trying to solve a particular motion planning problem. After a run is completed, several attributes are collected such as the number of generated states (\colname{graph\_states}), duration of the run (\colname{time}), length of the solution path (\colname{solution\_length}), clearance along the solution path (\colname{solution\_clearance}), etc. By default solutions are simplified (through a combination of short-cutting and smoothing), which usually significantly improves the solution quality at minimal time cost. Runs can terminate for a variety of reasons: a solution was found, the planner timed out (without any solution or with an approximate solution), or the planner crashed. We use an enumerate type for this attribute (stored in \colname{status}), and the labels for each value are stored in the \tbname{enums} table (not shown in Figure~\ref{fig:dbschema}).

\enlargethispage*{.4\baselineskip}
The \tbname{progress} table stores information periodically collected \emph{during} a run. This is done in a separate thread so as to minimize the effect on the run itself. Progress information is currently only available for optimizing planners. It is used to store the cost of the solution found at a particular time. By aggregating progress information from many runs for each planner, we can compare rates of convergence to optimality (see next section).

The database schema has been designed with extensibility in mind. Large parts of the schema are optional and other columns can be easily added. This does \emph{not} require new parsers or additional code. Instead, the log files contain enough structure to allow planners to define their own run and progress properties. Thus, when new log files are added to a database, new columns are automatically added to \colname{runs} and \colname{progress}. Planners that do not report on certain properties will just store ``N/A'' values in the corresponding columns. Additional run properties for a new type of planner are easily defined by storing key-value pairs in a dictionary of planner data which is obtained after each run. Additional progress properties are defined by adding a function to a list of callback functions.

Log files have a fairly straightforward plain text format that is easy to generate and parse\footnote{The complete syntax is specified at \url{http://ompl.kavrakilab.org/benchmark.html}.}. This makes it easy for other motion planning libraries to generate compatible log files, which can then be added to the same type of benchmark database. For example, MoveIt!'s benchmarking capabilities do not directly build on OMPL's benchmark capabilities, yet it can produce compatible benchmark log files. This makes it possible to see how a planning algorithm's performance changes when moving from abstract benchmark problems in OMPL to elaborate real-world settings created with MoveIt! (possibly from experimental data).

\section{Interactive analysis of results}

There are many different ways to visualize benchmark performance. It is nearly impossible to create a tool that can automatically select the ``right'' visualizations for a given benchmark database. We have therefore created a web site called Planner Arena (\url{http://plannerarena.org}), where benchmark data can be uploaded and selected results can be visualized. The web site interface is dynamically constructed based on the contents of the benchmark database: selection widgets are created automatically for the benchmark problems, the performance attributes, the planning algorithms, etc.
The code that powers Planner Arena is included in the OMPL distribution and can be run locally to evaluate one's own results privately or be modified to create custom visualizations.
There are currently three types of plots included on the Planner Arena site: overall performance plots, progress plots, and regression plots. We will describe these plots in more detail below.

\begin{figure*}
  \subfloat[][\textbf{Performance plot:} empirical cumulative distribution function of solution times for a rigid body benchmark]{\includegraphics[width=.48\linewidth]{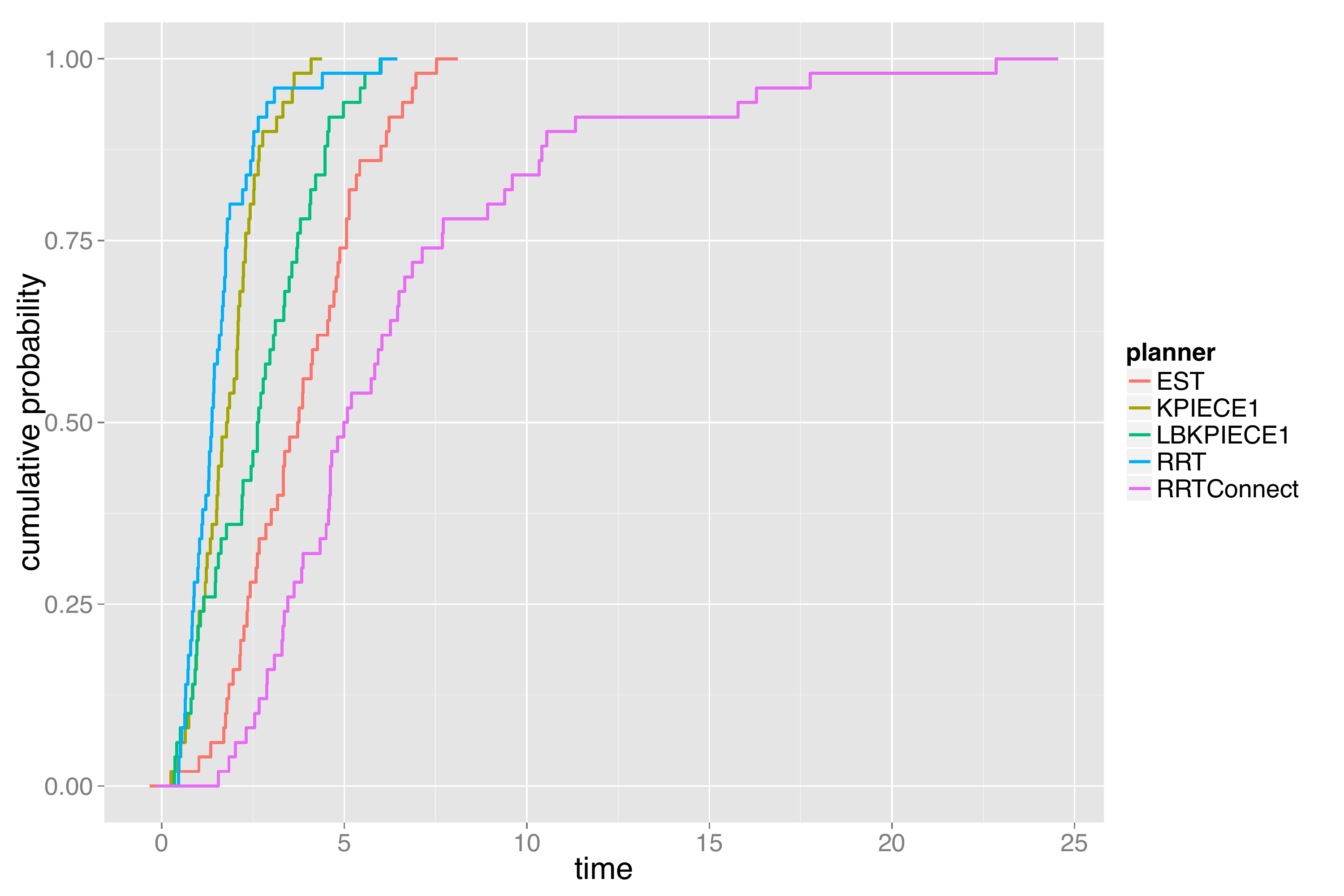}}\hfill
  \subfloat[][\textbf{Performance plot:} distance between best found approximate solution and goal for a kinodynamic problem]{\includegraphics[width=.45\linewidth]{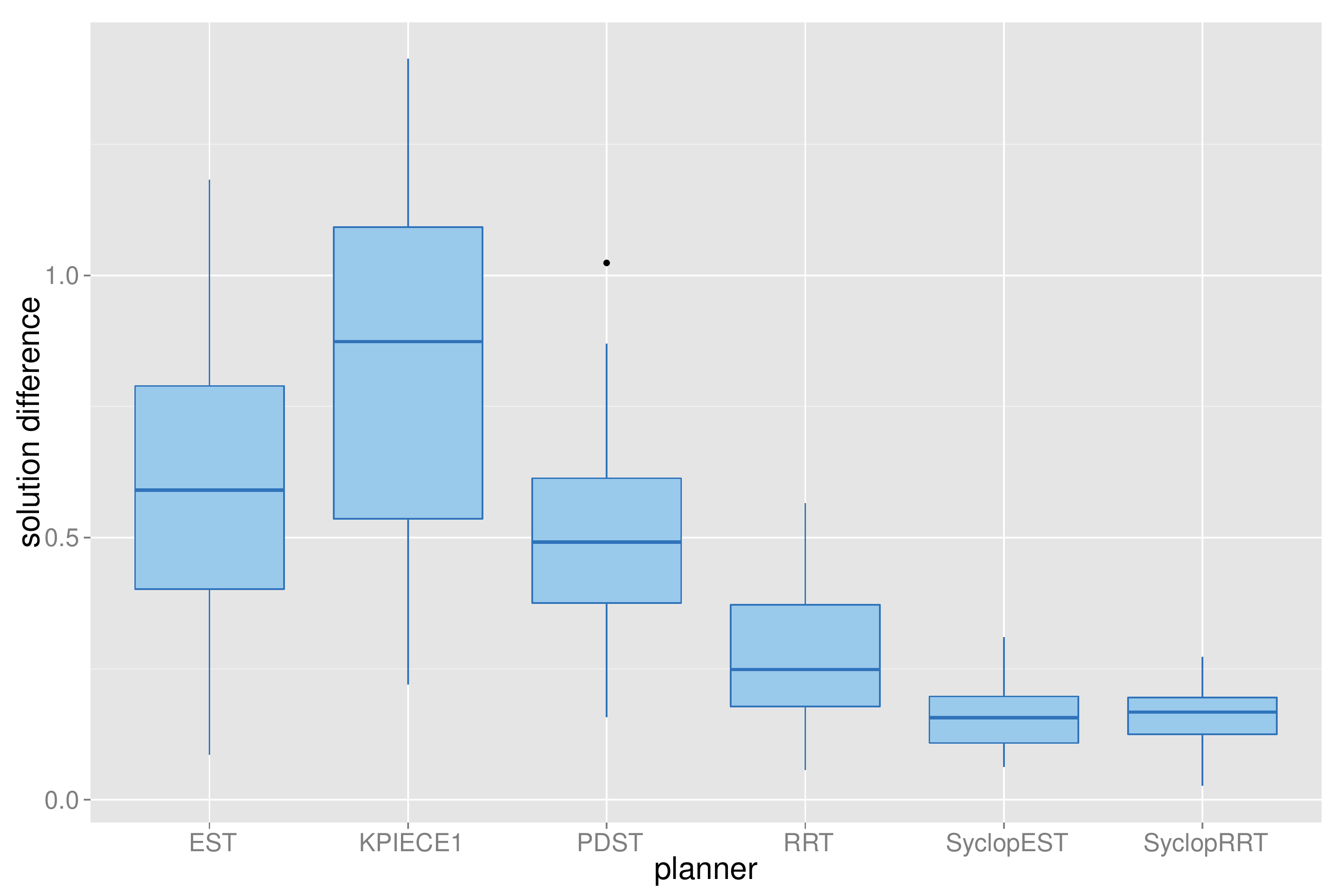}}\\
  \subfloat[][\textbf{Progress plot:} convergence rate of asymptotically optimal planners]{\includegraphics[width=.48\linewidth]{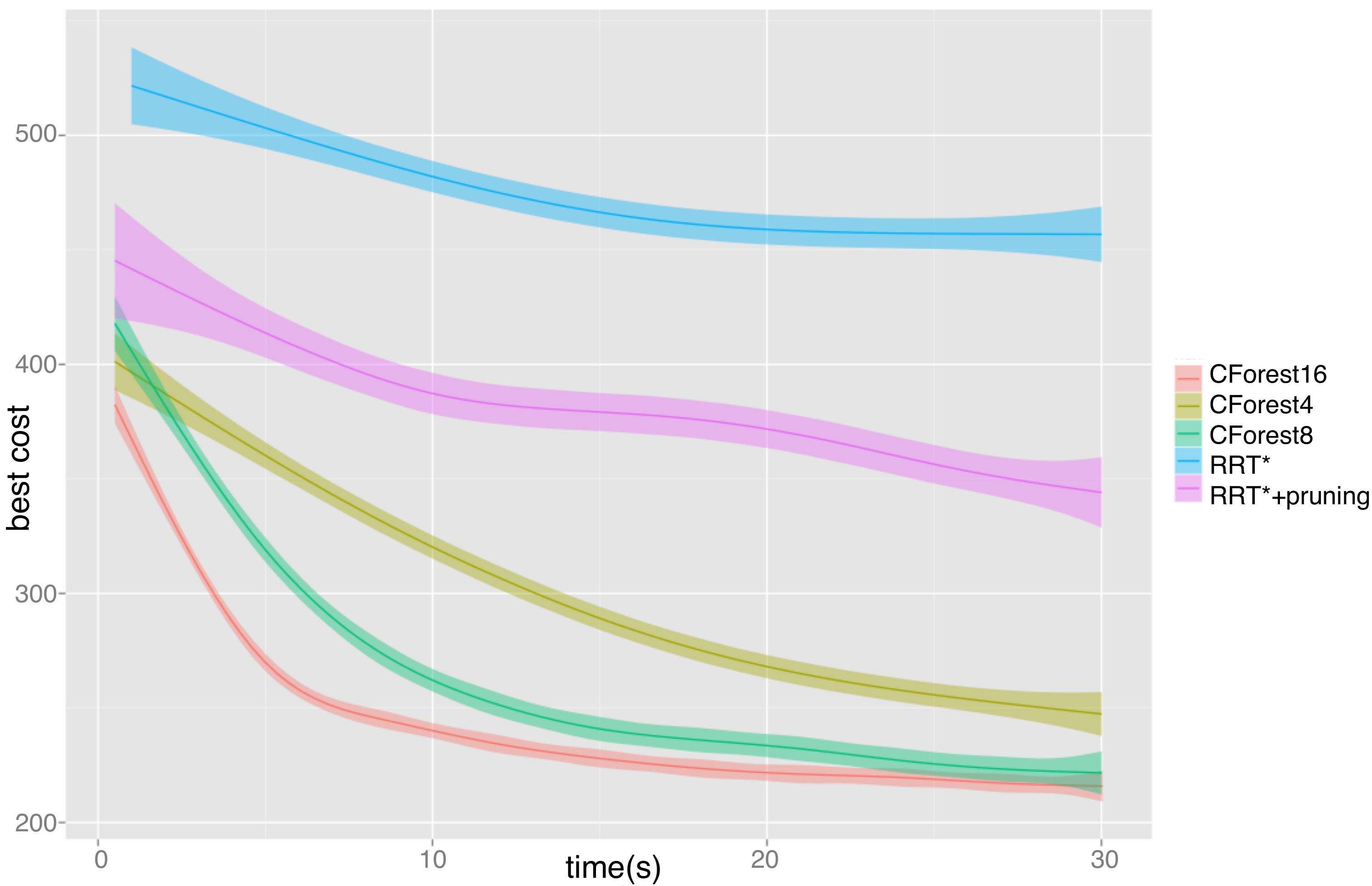}}\hfill
  \subfloat[][\textbf{Regression plot:} test results for a trivial benchmark]{\includegraphics[width=.48\linewidth]{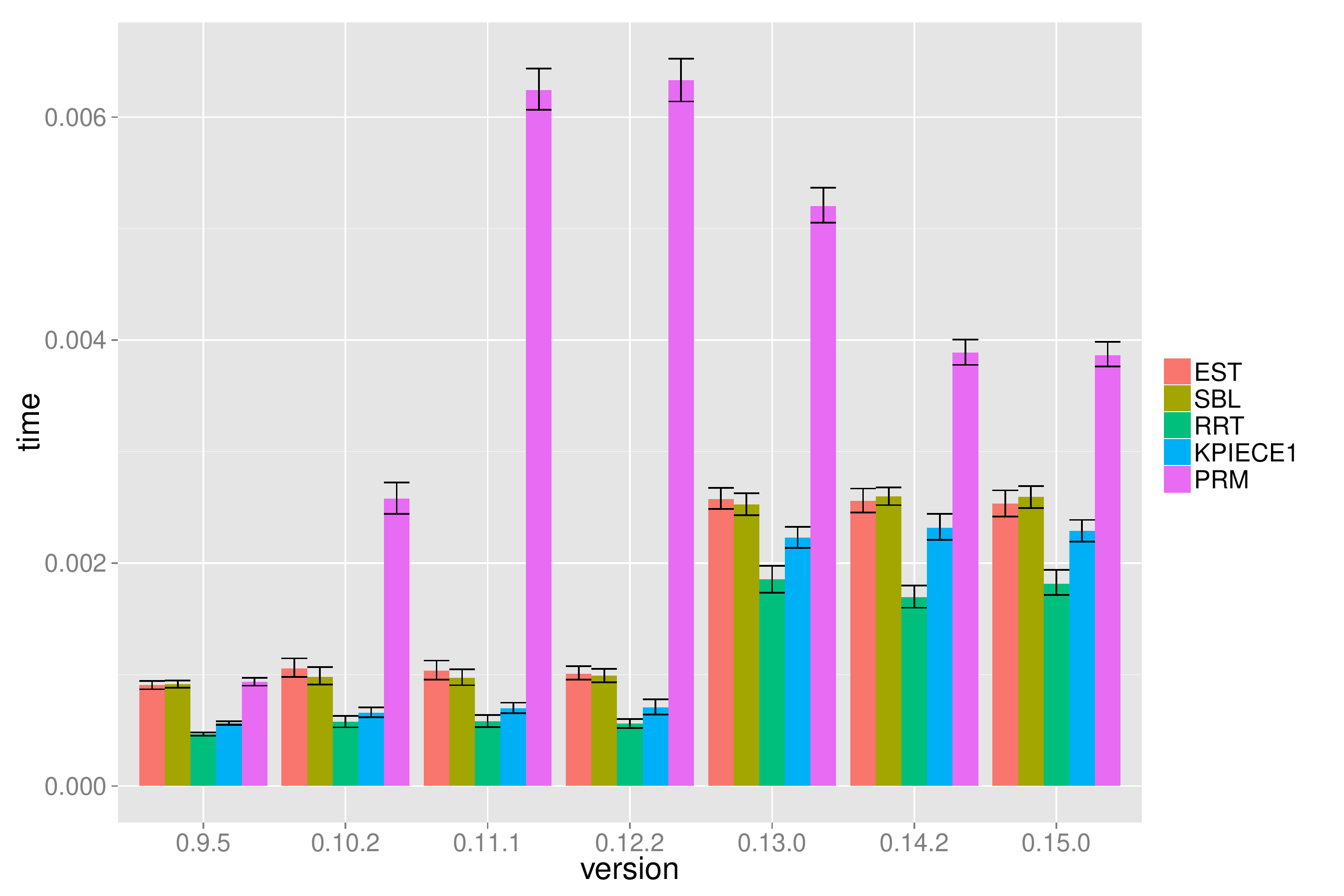}}
  \caption{Sample output produced from a benchmark database by the Planner Arena server.}
  \label{fig:plannerarena}
\end{figure*}

\paragraph*{Plots of overall performance}
The overall performance plots can show how different planners compare on various measures. The most common performance measure is the time it took a planner to find a feasible solution. By default, integer- and real-valued performance metrics (such as solution time) are plotted as box plots which provide useful summary statistics for each planner: median, confidence intervals, and outliers. However, in some cases visualizing the cumulative distribution function can reveal additional useful information. For instance, from Figure~\ref{fig:plannerarena}(a) one can easily read off the probability that a given planner can solve a particular benchmark within a specified amount of time.
For very hard problems where most planners time out without finding a solution, it might be informative to look at \emph{solution difference}: the gap between the best found solution and the goal (Figure~\ref{fig:plannerarena}(b)). For optimizing planners, it is often more interesting to look at the best solution found within some time limit. The overall performance page allows you to select a motion planning problem that was benchmarked, a particular benchmark attribute to plot, the OMPL version (in case the database contains data for multiple versions), and the planners to compare.

Most of the measures are plotted as box plots. Missing data is ignored. This is \emph{very} important to keep in mind: if a planner failed to solve a problem 99 times out of a 100 runs, then the average solution length is determined by one run! To make missing data more apparent, a table below the plot shows how many data points there were for each planner and how many of those were missing values.

Performance is often hard to judge by one metric alone. Depending on the application, a combination of metrics is often necessary to be able to choose an appropriate planner. For example, in our experience LBKPIECE (one of the planning algorithms in OMPL) tends to be among the fastest planners, but it also tends to produce longer paths. For time-critical applications this may be acceptable, but for applications that place greater importance on \emph{short} paths another planner might be more appropriate. There will also be exceptions to general trends. As another example, bidirectional planners (such as RRT-Connect) tend to be faster than unidirectional planners (such as RRT), but Figure~\ref{fig:plannerarena}(a) shows that this not always the case. This underscores the need for a good set of benchmark problems that are representative of different applications.

\paragraph*{Progress plots}
Some planners in OMPL are not limited to reporting information \emph{after} a run is completed, but can also periodically report information \emph{during} a run. In particular, for asymptotically optimal planners it is interesting to look at the convergence rate of the best path cost (e.g., path length).
By default, Planner Arena will plot the smoothed mean as well as a 95\% confidence interval for the mean (Figure~\ref{fig:plannerarena}(c)).
Optionally, individual measurements can  be shown as semi-transparent dots, which can be useful to get a better idea of the overall distribution.
Analogous to the performance plots, missing data is ignored. During the first couple seconds of a run, a planner may never find a solution path. Below the progress plot, we therefore plot the number of data points available for a particular planner at each 1 second time interval.

\begin{figure}
  \centering\includegraphics[width=.7\linewidth]{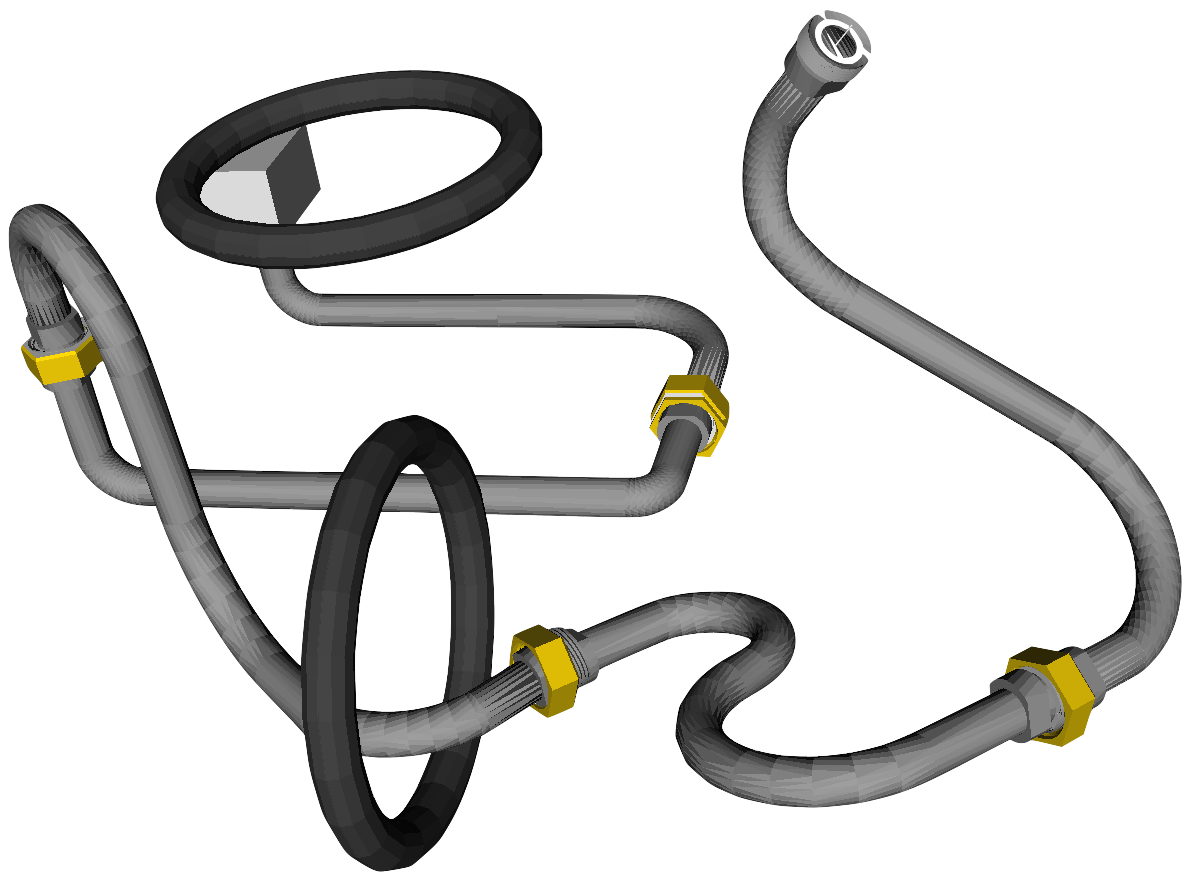}
  \includegraphics[width=.7\linewidth]{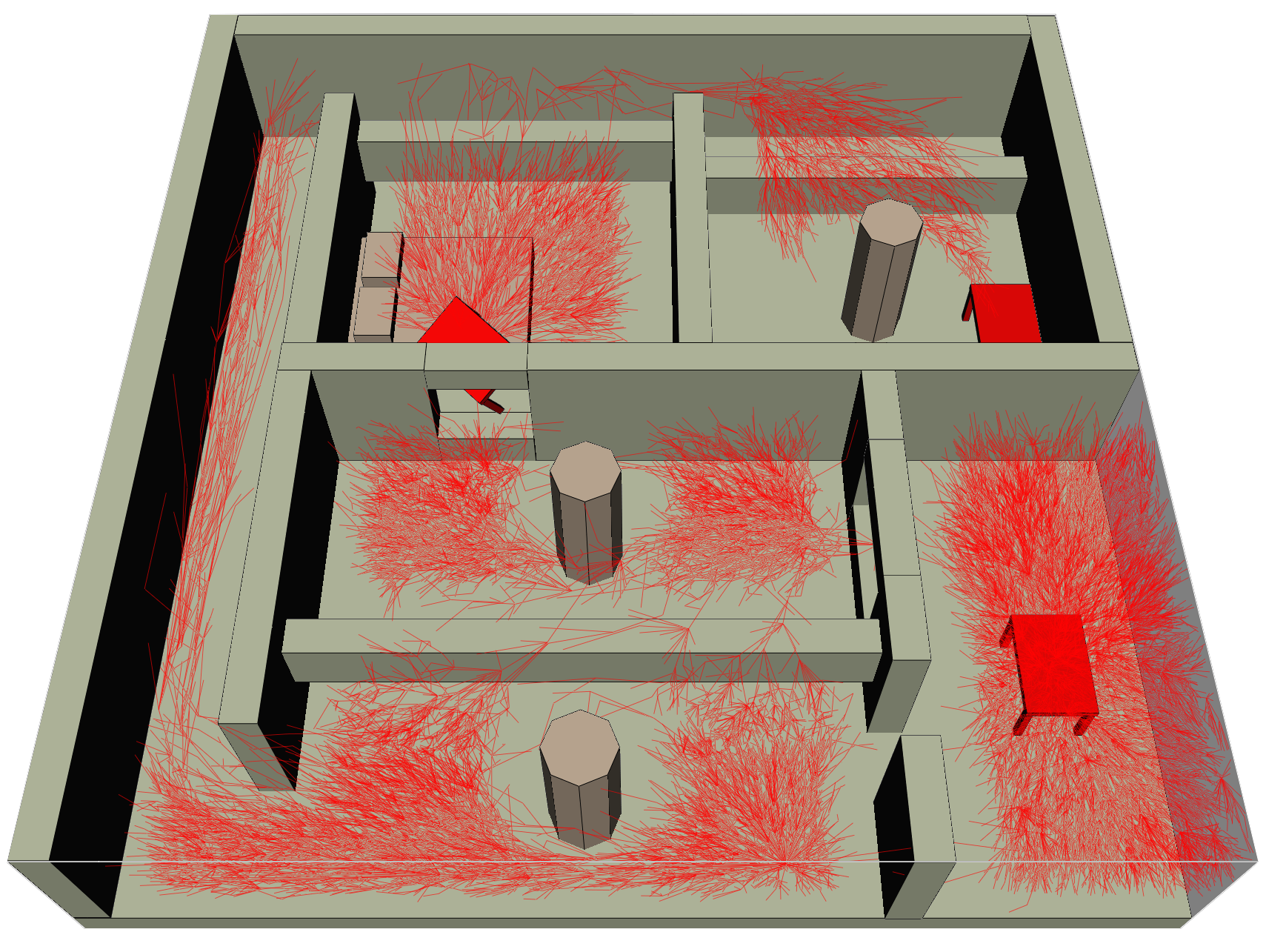}
  \caption{Two of the sample benchmark problems included on Planner Arena: one with a long twisty narrow passage (top) and one with several suboptimal decoy homotopy classes of paths (bottom).}
  \label{fig:plannerarena-benchmarks}
\end{figure}
\paragraph*{Regression plots}
Regression plots show how the performance of the same planners change over different versions of OMPL (Figure~\ref{fig:plannerarena}(d)). This is mostly a tool for developers using OMPL that can help in the identification of changes with unintended side-effects on performance. However, it also allows a user to easily compare the performance of a user's modifications to the planners in OMPL with the latest official release.
In regression plots, the results are shown as a bar plot with error bars.

Any of the plots can be downloaded in two formats: PDF and RData. The PDF format is useful if the plot is more or less ``camera-ready'' and might just need some touch ups. The RData file contains both the plot as well as all the data shown in the plot and can be loaded into R. The plot can be completely customized, further analysis can be applied to the data, or the data can be plotted in an entirely different way.

The default benchmark database stored on the server currently contains results for nine different benchmark problems. They include simple rigid body type problems, but also hard problems specifically designed for optimizing planners (problems that contain several suboptimal decoy homotopy classes), kinodynamic problems, and a multi-robot problem (see Figure~\ref{fig:plannerarena-benchmarks}).

\section{Discussion}


We expect that with input from leaders in the motion planning community as well as extensive simulations and experiments we can create a suite of motion planning benchmarks. We plan to develop benchmarks along two different directions. First, there are ``toy problems'' that isolate one of a number of common difficulties that could trip up a motion planning algorithm (such as a very narrow passage or the existence of many false leads). Such benchmarks may provide some insights that lead to algorithmic improvements. Second, we would like to develop a benchmark suite where performance (by some measure) is \emph{predictive} of performance of more complex real-world scenarios.

Other planning libraries can use the same set of benchmark problems. While OMPL could be extended with other planning algorithms, we recognize that for community-wide adoption of benchmarks it is important to adopt standard input and output file formats. The log file format and database schema for storing benchmark results described in this paper are general enough that they can be adapted by other motion planning software. This would allow for a direct comparison of different implementations of planning algorithms.

The Planner Arena web site makes it easy to interactively explore benchmark results.
At this point, we do not claim that the benchmarks included in the default database on Planner Arena form some sort of ``standard'' benchmark set, although they are representative of the types of problems that have been used in prior work \cite{amatobenchmarkpage}. Furthermore, the set of problems we present results for will increase over time.

\section*{Acknowledgments}

MM and LEK are supported in part by NSF NRI 1317849. The authors wish to thank Luis Torres for his contributions to the benchmarking capabilities in OMPL.


\bibliographystyle{IEEEtran}
\bibliography{mmoll-references}

\vspace*{\baselineskip}
\emph{\textbf{Mark Moll,}} Department of Computer Science, Rice University, Houston, TX. mmoll@rice.edu\\[.5\baselineskip]
\emph{\textbf{Ioan A. \c{S}ucan,}} Google[X], Mountain View, CA. isucan@gmail.com\\[.5\baselineskip]
\emph{\textbf{Lydia E.\ Kavraki,}} Department of Computer Science, Rice University, Houston, TX. kavraki@rice.edu

\end{document}